\documentclass[conference]{IEEEtran}
\IEEEoverridecommandlockouts
\usepackage{cite}
\usepackage{amsmath,amssymb,amsfonts}
\usepackage{algorithmic}
\usepackage{array}
\usepackage{graphicx}
\usepackage{multirow}
\usepackage{setspace}
\usepackage{textcomp}
\usepackage{xcolor}
\usepackage{url}
\def\BibTeX{{\rm B\kern-.05em{\sc i\kern-.025em b}\kern-.08em
    T\kern-.1667em\lower.7ex\hbox{E}\kern-.125emX}}
\begin{document}

\title{360ORB-SLAM: A Visual SLAM System for Panoramic Images with Depth Completion Network\\
\thanks{This research was supported by the National Natural Science Foundation of China under Grant 62202137, 62306097, China Postdoctoral Science Foundation under Grant 2023M730599, Zhejiang Provincial Natural Science Foundation of China under Grant LQ22F030004, Zhejiang Medical Electronics and Digital Health Key Laboratory under Grant MEDC202306, the Fundamental Research Funds for the Provincial Universities of Zhejiang GK239909299001-019, and the Research Foundation of Hangzhou Dianzi University (KYS335622091; KYH333122029M). 
}
}

\author{
\IEEEauthorblockN{Yichen Chen\textsuperscript{1}, Yuqi Pan\textsuperscript{1}, Ruyu Liu\textsuperscript{1, 2, *}, Haoyu Zhang\textsuperscript{1}, Guodao Zhang\textsuperscript{3}, Bo Sun\textsuperscript{2}, Jianhua Zhang\textsuperscript{4}}
\small
\textsuperscript{1}School of Information Science and Technology, Hangzhou Normal University, Hangzhou, 311121, China \\
\textsuperscript{2}Haixi Institutes, Chinese Academy of Sciences Quanzhou Institute of Equipment Manufacturing, Quanzhou, 362000, China \\
\textsuperscript{3}Department of Digital Media Technology, Hangzhou Dianzi University, Hangzhou, 310018, China \\
\textsuperscript{4}School of Computer Science and Engineering, Tianjin University of Technology, Tianjin, 300384, China \\
\textsuperscript{*}Corresponding author. Email address: \url{lry@hznu.edu.cn}
\normalsize
}

\maketitle

\begin{abstract}
To enhance the performance and effect of AR/VR applications and visual assistance and inspection systems, visual simultaneous localization and mapping (vSLAM) is a fundamental task in computer vision and robotics. However, traditional vSLAM systems are limited by the camera's narrow field-of-view, resulting in challenges such as sparse feature distribution and lack of dense depth information. To overcome these limitations, this paper proposes a 360ORB-SLAM system for panoramic images that combines with a depth completion network. The system extracts feature points from the panoramic image, utilizes a panoramic triangulation module to generate sparse depth information, and employs a depth completion network to obtain a dense panoramic depth map. Experimental results on our novel panoramic dataset constructed based on Carla demonstrate that the proposed method achieves superior scale accuracy compared to existing monocular SLAM methods and effectively addresses the challenges of feature association and scale ambiguity. The integration of the depth completion network enhances system stability and mitigates the impact of dynamic elements on SLAM performance.
\end{abstract}


\section{Introduction}
 Simultaneous Localization and Mapping (SLAM) \cite{b1} refers to the process of constructing an environmental model and simultaneously estimating the agent's motion while being equipped with specific sensors, without prior knowledge of the environment. As AR/VR applications and visual assistance and inspection systems continue to evolve and become more prevalent, real-time and robust SLAM systems play a crucial role in various robotic applications such as autonomous driving, robot navigation, and augmented reality.

Monocular visual SLAM (vSLAM) systems cannot directly measure depth information and rely on motion estimation and triangulation, leading to scale uncertainty. In addition, the performance of vSLAM is significantly influenced by the camera's field of view (FOV). A larger FOV captures more visual information, enabling better landmark tracking and increased frame overlap. This is crucial for robust camera pose estimation \cite{b2}, especially in sparsely featured environments. Wide FOV cameras also minimize the interference of dynamic elements, allowing effective removal of outliers and accurate pose estimation even in dynamic environments.
\begin{figure}
    \centering
    \includegraphics[width=0.8\linewidth]{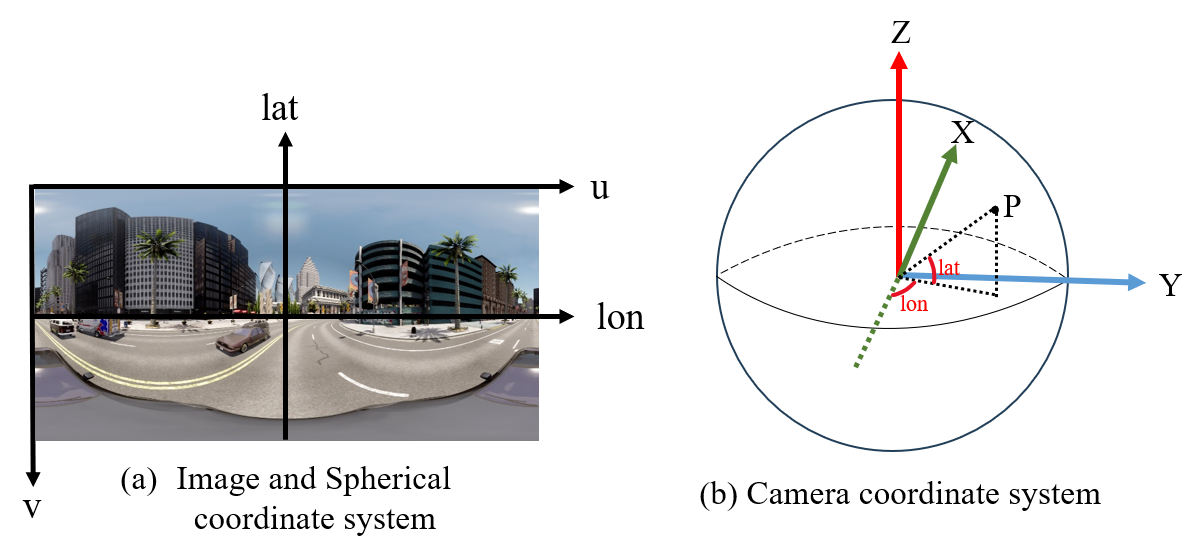}
    \caption{Coordinate systems}
    \label{fig:coordinate}
\end{figure}

However, when the camera's FOV exceeds 180 degrees, nonlinear distortions occur, making it challenging to rectify the images into perspective images. Conventional pinhole camera models used in classical SLAM systems, such as ORB-SLAM2 \cite{b3}, DSO \cite{b4}, and LSD-SLAM \cite{b5}, struggle with larger FOV images. Recent advancements, like Open-vSLAM \cite{b6} and omnidirectional DSO \cite{b7}, have adapted ORB-SLAM and DSO to fisheye and catadioptric cameras. ORB-SLAM3 \cite{b8} introduced the Kannala Brandt camera model \cite{b9} to handle FOV exceeding 180 degrees. However, none of these systems can handle panoramic cameras with a 360-degree FOV. Recently, 360VO \cite{b10} improved the epipolar constraints based on DSO to handle panoramic cameras with a 360-degree FOV. However, DSO is more sensitive to changes in lighting conditions and performs less effectively than feature-based SLAM systems in unstable lighting environments.

Against the backdrop of the flourishing development of artificial intelligence, deep learning techniques are gradually being applied in vSLAM. The integration of deep learning into SLAM serves to enhance its generalizability to new environments and improve the accuracy and robustness of SLAM algorithms while reducing computational costs. Panoramic camera-based SLAM systems are particularly concerned with a 360-degree global environment, offering a broader FOV, necessitating more precise depth information about the surroundings. Hence, we employ depth completion techniques based on confidence propagation to densify the sparse depth input, resulting in robust dense panoramic depth information for subsequent localization and mapping tasks.
\begin{figure}
    \centering
    \includegraphics[width=0.8\linewidth]{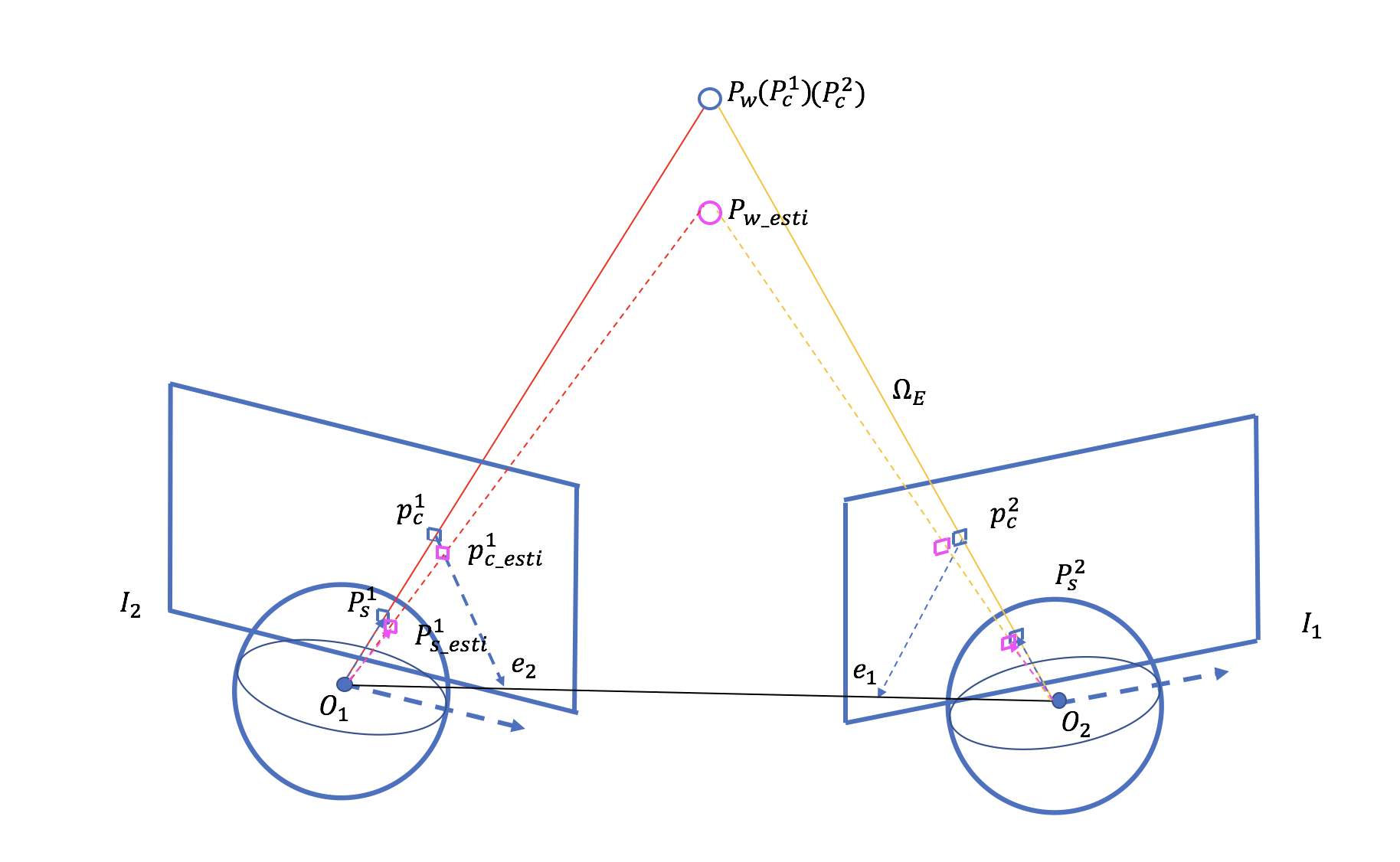}
    \caption{Triangulation based on a panoramic camera model}
    \label{fig:panoramic_Tri}
\end{figure}
\begin{figure}
    \centering
    \includegraphics[width=0.8\linewidth]{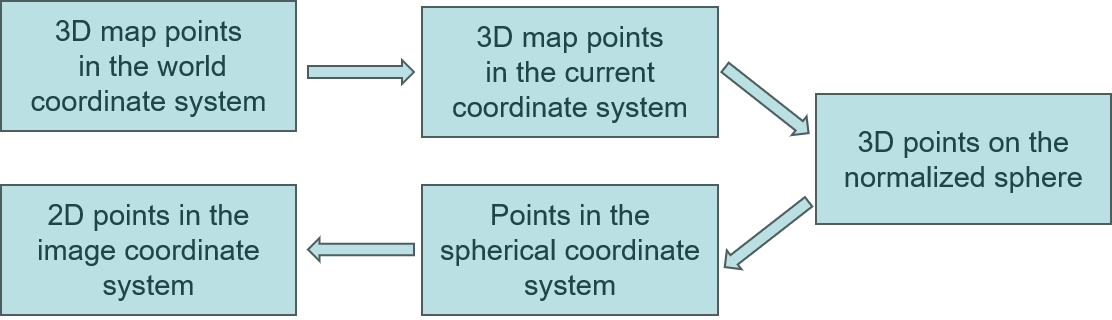}
    \caption{Panoramic projection flow chart}
    \label{fig:panoramic_projection_process}
\end{figure}

ORB-SLAM2 is a modern SLAM system that balances excellent efficiency and accuracy. It efficiently extracts features in real-time while maintaining strong rotational invariance. Therefore, this paper proposes a vSLAM system for panoramic cameras called 360ORB-SLAM, based on the ORB-SLAM2 framework. The system incorporates a panoramic triangulation module designed to generate sparse depth information from ORB features, which is then fed into a depth completion network for reconstructing dense depth maps. Finally, the three main threads of the ORB-SLAM2 system, namely tracking, local mapping, and loop closing, are harnessed to achieve localization and mapping functions for the panoramic camera.

The main contributions and innovations of this paper are as follows:
\begin{enumerate}
  \item A panoramic triangulation module is designed to generate sparse depth information from ORB features. This module considers the specific characteristics of panoramic cameras and incorporates spherical projection models and epipolar geometry constraints, effectively improving the accuracy and stability of map points.
  \item A depth completion network is introduced to extract potential geometric features and generate dense depth maps from sparse depth information. This network utilizes a multi-scale confidence propagation network and a multimodal fusion network, using RGB images and confidence as guidance, to achieve high-quality depth estimation.
  \item The dense depth maps recovered from the deep completion network are applied in the SLAM system, to address the scale drift issue of monocular cameras and improve pose estimation accuracy. Additionally, omnidirectional perception is utilized to capture more environmental information and reduce interference from dynamic points, thereby enhancing the robustness of visual odometry.
  \item A panoramic dataset is constructed for intelligent driving applications, and comparative experiments are conducted on this dataset. The results demonstrate significant advantages of the proposed system over ORB-SLAM2 in terms of absolute trajectory error and scale factor, without encountering tracking failures.
\end{enumerate}

\section{Camera Model}
The camera model is a mathematical representation that describes the mapping relationship between 3D points in space and their projection onto a 2D image plane. The camera model employed in this paper is the spherical projection model, which is a method for converting panoramic images into planar images. To describe the position and transformations of a point in the spherical projection model, this paper utilizes three different coordinate systems, as illustrated in Fig. \ref{fig:coordinate}. These are the image coordinate system, the spherical coordinate system, and the camera coordinate system.

Conventionally, the camera space is denoted as $\Omega$, while the corresponding image space is represented as $\Psi$.  The image size is expressed as $[H, W]^T$. We referenced and modified the camera model of 360VO \cite{b10} to adapt it to our system. The specific details of the camera model are as follows:
\begin{equation}
\label{eq:intrinsic_matrix}
K = 
\begin{bmatrix}
    f_x & 0 & c_x \\
    0 & f_y & c_y
\end{bmatrix}
=
\begin{bmatrix}
    W/2\pi & 0 & W/2 \\
    0 & -H/\pi & H/2
\end{bmatrix}
\end{equation}
\begin{equation}
\label{eq:projection}
\pi(\boldsymbol{P}) = 
\begin{bmatrix}
    u \\
    v 
\end{bmatrix}
=
K
\begin{bmatrix}
    lon \\
    lat 
\end{bmatrix}
=
K
\begin{bmatrix}
    arctan(-Y_c/X_c) \\
    arcsin(Z_c/r) 
\end{bmatrix}
\end{equation}
\begin{equation}
\label{eq:inverse_projection}
\pi^{-1}(\boldsymbol{u}, d) = \boldsymbol{P} = 
\begin{bmatrix}
    X \\ Y \\ Z
\end{bmatrix}
=
d
\begin{bmatrix}
    -cos(lat) \cdot cos(lon) \\
    cos(lat) \cdot sin(lon) \\
    sin(lat)
\end{bmatrix}
\end{equation}
\begin{equation}
\label{eq:lon_lat}
\begin{bmatrix}
    lon \\ lat
\end{bmatrix}
=
K^{-1}
\begin{bmatrix}
    u \\ v
\end{bmatrix}
=
\begin{bmatrix}
    f_x^{-1} & 0 & -f_x^{-1}c_x \\
    0 & f_y^{-1} & -f_y^{-1}c_y
\end{bmatrix}
\begin{bmatrix}
    u \\ v
\end{bmatrix}
\end{equation}
\begin{figure*}
    \centering
    \includegraphics[width=1\textwidth]{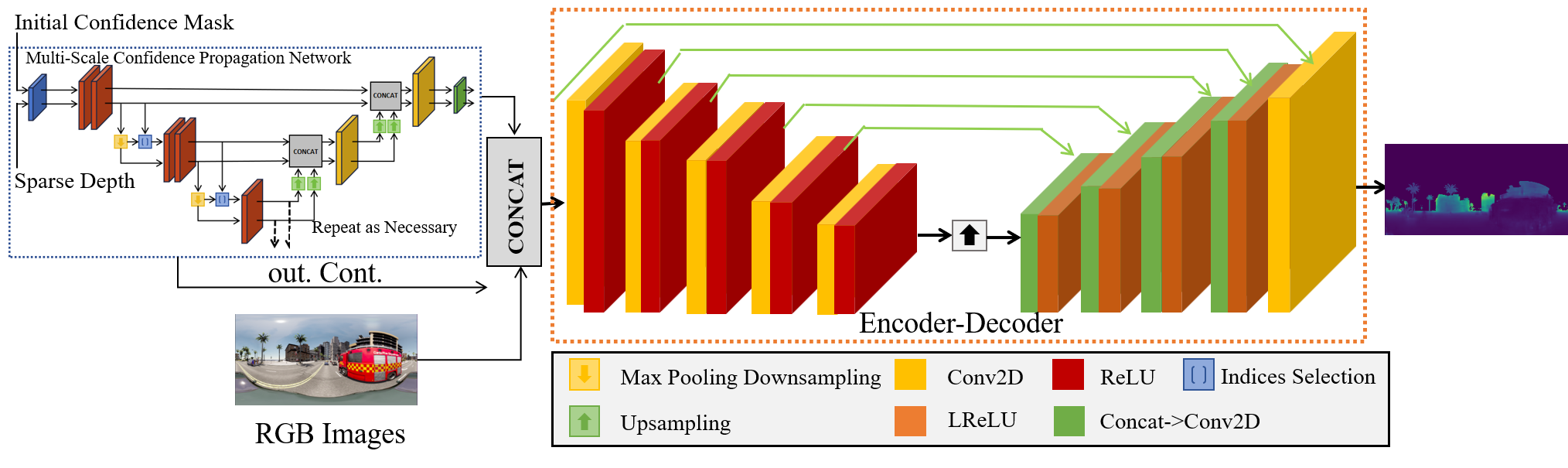}
    \caption{Deep completion network framework diagram}
    \label{fig:network}
\end{figure*}
$K$ is the camera's intrinsic matrix, and $f_x$, $f_y$, $c_x$, $c_y$ are specific intrinsic parameters that are derived from $H$ and $W$. Equation \eqref{eq:projection} and \eqref{eq:inverse_projection}  respectively represent the projection function and the up-projection function, describing the transformation process between a 3D point $\boldsymbol{P}$ in camera space and its projection $\boldsymbol{u}$ onto the 2D pixel coordinates of the image plane. $lon$ and $lat$ represent the longitude and latitude respectively in the spherical coordinate system.
\section{The Proposed Method}

\subsection{Panoramic Triangulation Module}
After receiving keyframes selected by the visual front-end, the system inserts these keyframes into the map and, through triangulation, reconstructs the map points, thereby building the environment map. As shown in Fig.~\ref{fig:panoramic_Tri}, the algorithm for map point recovery based on a panoramic camera is as follows:

Firstly, the system identifies the neighboring keyframe $I_1$, with the highest level of visual overlap with the current frame $I_2$. It calculates the disparity between $I_2$ and $I_1$ to determine if there is sufficient parallax. Only frames that meet the criteria of having significant disparity are eligible for panoramic triangulation of the 3D points. Subsequently, by minimizing the reprojection error, the system estimates the pose and calculates the essential matrix $E_{12}$ between the two keyframes. The essential matrix $E_{12}$ is then utilized to find matching features between the $I_2$ and $I_1$. These matching features must satisfy certain epipolar geometry constraints: the angle between the feature point $P_s^2$ on the spherical surface in the $I_2$'s coordinate system and the epipolar $e_1$ should not be too small, meeting a specific threshold $|\overrightarrow{O_2e_1} \cdot \overrightarrow{O_2P_s^2}| > \mu$. Additionally, the distance $d_{P_s^2 \rightarrow \Omega_E}$ between $P_s^2$ and the epipolar plane $\Omega_E$ should be smaller than a certain threshold. In the context of the panoramic imaging model, the specific constraints are as follows: 
\begin{equation}
\label{equ:Omega_E}
\Omega_E = aX + bY + cZ
\end{equation}
\begin{equation}
\label{equ:[abc]}
\begin{bmatrix}
    a & b & c
\end{bmatrix}_{\Omega_E}^T = P_s^2E_{12}
\end{equation}
\begin{equation}
\label{equ:d_p_Omega}
d_{P_s^2 \rightarrow \Omega_E} = \frac{(aX_s + bY_s + cZ_s)^2}{a^2 + b^2 + c^2}
\end{equation}

After identifying the corresponding features using epipolar geometry constraints, the next step is to estimate the depth of these features in the panoramic sphere model. Assuming the 3D points in the world coordinate system are denoted as $P_w$, and the poses of $I_1$ and $I_2$ are represented by $T_{c_1w}$ and $T_{c_2w}$, respectively. The specific method is as follows:
\begin{equation}
\label{equ:sp_TP}
\left\{
\begin{aligned}
s_{c_1}p_{c_1} &= T_{c_1w}P_w \\ 
s_{c_2}p_{c_2} &= T_{c_2w}P_w
\end{aligned}
\right.
\end{equation}
\begin{equation}
\label{equ:pT}
\begin{bmatrix}
    p_{c_1} \times T_{c_1w} \\
    p_{c_2} \times T_{c_2w}
\end{bmatrix}
P_w = Ax = 0
\end{equation}

By performing SVD of the matrix $A = uWv^T$, we can obtain the estimated coordinates of the map point as $P_{w\_e}$. Verify the estimated map points for their visibility in front of the camera, i.e., ensuring that the depth values of map points are positive. Subsequently, reproject the map points onto the panoramic sphere of $I_2$ and $I_1$, calculating the angular deviation between the reprojected feature point $P_{s\_e}(X_{s\_e}, Y_{s\_e}, Z_{s\_e})$ and the original feature point $P_s(X_s, Y_s, Z_s)$, ensuring that the angular error falls within a specified threshold range, specified as:
\begin{equation}
\label{eq:re_or_angle}
\cos\langle \overrightarrow{OP_{s\_e}}, \overrightarrow{OP_s}\rangle = \frac{X_{s\_e}X_s + Y_{s\_e}Y_s + Z_{s\_e}Z_s}{\sqrt{X_{s\_e}^2 + Y_{s\_e}^2 + Z_{s\_e}^2} \cdot \sqrt{X_s^2 + Y_s^2 + Z_2^2}}
\end{equation}

Finally, utilize the panoramic projection model to project the map points of the current frame onto the coordinate system of neighboring keyframes. Determine if there are any duplicate map points and perform fusion accordingly.
\subsection{Sparse Depth Generation Module}

To obtain accurate and reliable sparse depth information, for each frame, project the newly generated map points obtained from the panoramic triangulation module and all previously generated map points within the current frame's FOV onto the current frame through panoramic projection. This process results in a sparse depth map.

As shown in Fig.~\ref{fig:panoramic_projection_process}, the specific panoramic projection process begins with transforming the map points in the world coordinate system to the current coordinate system through pose transformation. Specifically,
\begin{equation}
\label{P_ji}
\boldsymbol{P_{ji}} = R_{wi}^{-1}\boldsymbol{P_{jw}} - R_{wi}^{-1}t_{wi}
\end{equation}
\begin{equation}
\label{uji}
\boldsymbol{u_{ji}'} = \pi(\boldsymbol{P_{ji}})
\end{equation}
where, $i$ and $j$ represent the index of frames, $w$ represents the world coordinate system. $\boldsymbol{u}$ represents pixel coordinates, and $\boldsymbol{P}$ represents the coordinates of a 3D map point. $R$ and $t$ denote the rotation matrix and translation vector, respectively. Projection points that fall outside the image should be discarded to avoid affecting the stability of the system.

By eliminating $d$ in \eqref{eq:inverse_projection} and solving for $lat$ and $lon$, substitute the values into equation \eqref{eq:projection} to obtain the image coordinates. Then, store the depth value $d$ at the corresponding coordinates in the sparse depth map.

\subsection{Dense Depth Completion Module}
SLAM, as a pixel-level visual task, requires careful consideration of the depth information for each map point to achieve accurate and reliable dense depth estimation. Obtaining precise and reliable dense depth is essential. To effectively leverage sparse data, we employ a fusion approach that combines sparse depth (SD) measurements with RGB images to estimate dense depth. However, a challenge arises due to the inconsistent data distribution between RGB and SD. Sparse and irregular depth measurements exhibit different density distributions compared to full-resolution RGB images. Therefore, we introduce a confidence-based depth completion network \cite{b14} that takes into account the sparsity characteristics and follows a strategy from sparse to dense, enabling robust dense depth estimation.

We utilize a multi-scale confidence propagation network to extract latent geometric features from sparse data and generate dense feature maps. Furthermore, using RGB images and the output confidences as guidance, we apply them to the output depth map of the multi-scale confidence prediction network. By employing a depth completion network with multi-modal fusion, we achieve accurate dense output.
\subsubsection{Multi-scale Confidence Propagation Network}
 The multi-scale confidence propagation network adopts an encoder-decoder structure, as illustrated in Fig.  \ref{fig:network}. In this architecture, the standard convolutional layers in the encoder-decoder framework are replaced with normalized convolution (NConv) layers.  The utilization of NConv enables the network to compute the confidence level for each layer's output and propagate this confidence to subsequent layers. This approach leverages the concept of confidence signals to effectively represent neighboring pixels, thereby enhancing the network's ability to diffuse and densify pixels. 

The NConv-based multi-scale module generates confidence features for each receptive field of the depth map, enhancing the network's ability to capture informative geometric features. This contributes to improved depth estimation performance.
\subsubsection{Depth completion network based on multimodal fusion}
The dense depth obtained solely from the multi-scale confidence propagation network has local weaknesses, exhibiting discontinuities in regions with edges and rough surfaces. To address this, this paper introduces a multimodal fusion network that utilizes RGB images with rich feature characteristics to guide spare depth completion, resulting in more accurate dense depth estimation. As shown in Fig. \ref{fig:network}, we first concatenate the depth data and confidence obtained from the multi-scale unsupervised network with the RGB image. This concatenated input is then fed into the encoder-decoder network. Through a series of downsampling and upsampling operations, the network progressively processes the input and generates dense depth output at the final layer.
\begin{figure}
    \centering
    \includegraphics[width=0.9\linewidth]{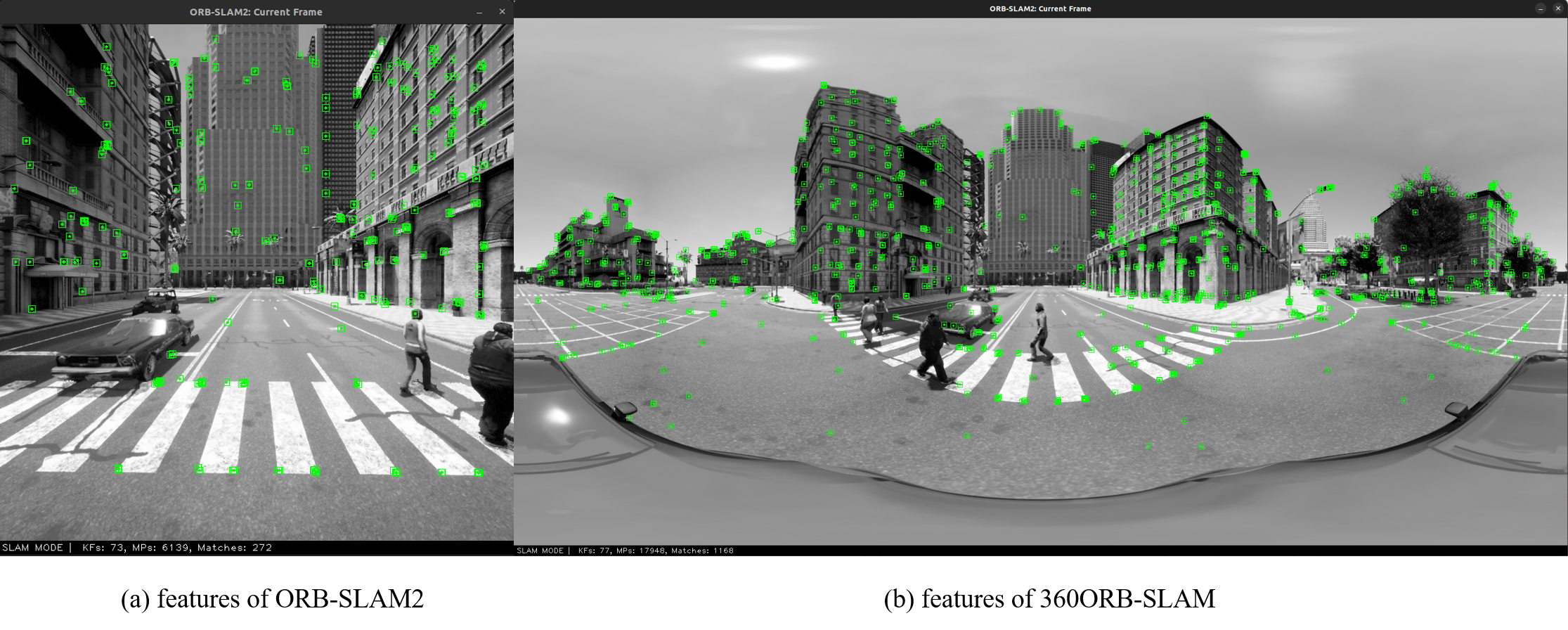}
    \caption{Comparison of monocular and panoramic feature detection}
    \label{fig:features}
\end{figure}
\begin{figure}
    \centering
    \includegraphics[width=0.7\linewidth]{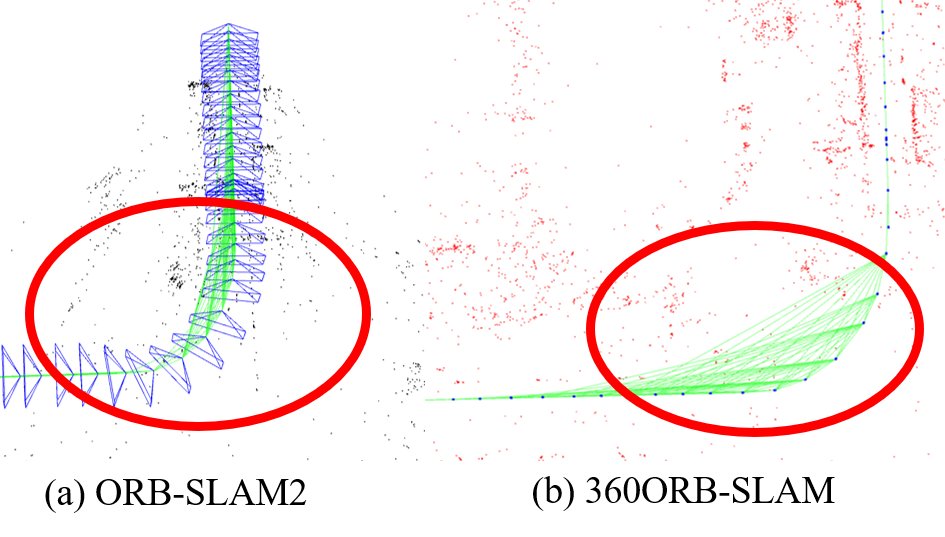}
    \caption{Comparison of monocular and panoramic covisibility graph}
    \label{fig:Map_compare}
\end{figure}

Considering that the sparse depth acquisition method in this paper is the ORB-SLAM algorithm, to better integrate with the SLAM system and utilize the advantages of feature points for more accurate and robust results, we use the sparse depths obtained from the earlier ORB feature point method as sparse inputs for network training.
\subsection{Dense Depth Application Module}
The dense depth map recovered by the depth completion module effectively addresses the scale drift issue in monocular cameras and further enhances the accuracy of pose estimation. Therefore, compared to classical monocular visual systems, in this paper, the dense depth map is utilized in modules such as initialization, tracking, and local mapping. The dense depth map provides depth information for map points, and with the aid of omnidirectional perception, more environmental information is acquired, reducing the presence of dynamic objects within the FOV. As a result, the stability and robustness of SLAM are improved.
\section{Experiments}
\subsection{Dataset}
To validate the effectiveness of the proposed system, a panoramic dataset was constructed, where each frame contains dense ground truth depth and ground truth pose. Specifically, in this paper, panoramic sequences were rendered using the realistic urban scene models provided in the Carla simulation environment \cite{b16}. The dataset sequences cover various times of the day, including morning, noon, and evening, capturing images under varied lighting conditions. The dataset allows for manual manipulation of the number of vehicles in the environment, ensuring diversity and randomness. It consists of $5$ sequences, with an average of $2000$ frames per sequence. The panoramic images have dimensions of $1024 \times 2048$ pixels and are created by stitching together six images of size $768 \times 768$ pixels taken from the front, back, left, right, top, and bottom perspectives.
\subsection{Experimental Setup}
The experimental platform for this paper is the Ubuntu 22.04 LTS operating system. For the SLAM part, the CPU used is an Intel Core i7-10750H @ 2.6 GHz, with GPU models NVIDIA GeForce GTX 1650 Ti and Intel UHD Graphics. For the deep learning part, the CPU used is an Intel Core i7-8700 @ 3.20 GHz, with a GPU model GeForce RTX 3090 having 24GB of VRAM. The deep learning framework used is version 1.13.1, and the CUDA version is 12.0.
\begin{table}
\centering
\label{table:Features}
\renewcommand{\arraystretch}{1.2}
\caption{Comparison of feature detection}
\resizebox{\columnwidth}{!}{
\begin{tabular}{ccccc}
\hline
\multirow{2}{*}{Sequences} & \multicolumn{2}{c}{\begin{tabular}[c]{@{}c@{}}Keyframe Features \\ ("X" means LOST)\end{tabular}} & \multirow{2}{*}{Frames} & \multirow{2}{*}{Time} \\ \cline{2-3}
                           & ORB-SLAM2                                     & 360ORB-SLAM                                       &                         &                       \\ \hline
Town1                      & 300                                           & \textbf{615}                                      & 1000                    & Noon                  \\
Town2                      & 265                                           & \textbf{761}                                      & 2000                    & Sunset                \\
Town3                      & X                                             & \textbf{1117}                                     & 2850                    & Morning               \\
Town4                      & 240                                           & \textbf{716}                                      & 3000                    & Noon                  \\
Town5                      & X                                             & \textbf{686}                                      & 1000                    & Sunset                \\ \hline
\multicolumn{3}{l}{*The bolded values represent the optimal values for each line.}
\end{tabular}
}
\renewcommand{\arraystretch}{1}
\end{table}
\subsection{Results and Analysis}
To validate the effectiveness and robustness of the proposed method in this study, due to the current lack of open-source panoramic SLAM systems, ORB-SLAM2 was chosen as the benchmark for horizontal comparison. Furthermore, for vertical comparison, the ATE and SF were compared with 360ORB-SLAM-Tri, which solely relies on triangulation for depth estimation. The proposed method is represented by 360ORB-SLAM-DC, incorporating a depth completion network. We utilizes the Scale Factor (SF) and Root Mean Squared Error (RMSE) of Absolute Trajectory Error (ATE) as evaluation metrics. The experiments were conducted on each sequence for $5$ runs in these three systems, and the median results of each run were listed to demonstrate the stochasticity of the multi-threaded systems.
\begin{table}
\label{table:RMSE}
\centering
\renewcommand{\arraystretch}{1.2}
\caption{Comparison of Root Mean Square Error (RMSE).} 
\resizebox{\columnwidth}{!}{
\begin{tabular}{ccccc}
\hline
\multirow{2}{*}{Sequences} & \multicolumn{3}{c}{\begin{tabular}[c]{@{}c@{}}Keyframe Trajectory (m)\\ ("X" means LOST)\end{tabular}} & \multirow{2}{*}{Frames} \\ \cline{2-4}
                           & ORB-SLAM2                        & 360ORB-SLAM-Tri                  & 360ORB-SLAM-DC                   &                         \\ \hline
Town1                      & 0.14886                          & \textbf{0.14782}                 & 0.16760                          & 1000                    \\
Town2                      & \textbf{0.14842}                 & 0.22466                          & 0.18319                          & 2000                    \\
Town3                      & X                                & 0.10128                          & \textbf{0.07288}                 & 2850                    \\
Town4                      & 0.87395                          & \textbf{0.05215}                 & 0.08820                          & 3000                    \\
Town5                      & X                                & 0.20135                          & \textbf{0.19273}                 & 1000                    \\ \hline
\multicolumn{3}{l}{*The bolded values represent the optimal values for each line. }
\end{tabular}
}
\renewcommand{\arraystretch}{1}
\end{table}
\subsubsection{Feature Detection and Matching}
A major advantage of using a panoramic camera is its ability to capture more features in the input images. As shown in Fig.\ref{fig:features}, our system demonstrates a significant increase in feature detection compared to ORB-SLAM2. On average, our system observes 2.83 times more features in a single frame. This increased feature observation is accompanied by a greater degree of image overlap between frames, leading to stronger constraints between keyframes in our system. This is illustrated by the red circles in the covisibility graph shown in Fig.\ref{fig:Map_compare}. Table I presents the quantitative results, highlighting the superior performance of our system in feature detection and matching compared to ORB-SLAM2. In contrast, ORB-SLAM2 exhibits a higher tracking failure rate, primarily due to low inter-keyframe covisibility. Such failures are triggered by rapid camera motion, changes in lighting conditions, or abrupt scene variations within the FOV. In contrast, 360ORB-SLAM using panoramic images shows enhanced stability and robustness in the experimental results, with no tracking failures.
\begin{table}
\centering
\label{table:SF}
\renewcommand{\arraystretch}{1.2}
\caption{Comparison of Scale Factor (SF)}
\resizebox{\columnwidth}{!}{
\begin{tabular}{ccccc}
\hline
\multirow{2}{*}{Sequences} & \multicolumn{3}{c}{\begin{tabular}[c]{@{}c@{}}Keyframe Scale Factor\\ ("X" means LOST)\end{tabular}} & \multirow{2}{*}{Frames} \\ \cline{2-4}
                           & ORB-SLAM2                   & 360ORB-SLAM-Tri                   & 360ORB-SLAM-DC                     &                         \\ \hline
Town1                      & 29.87543                    & 2.44838                           & \textbf{1.31180}                   & 1000                    \\
Town2                      & 30.11784                    & 19.60934                          & \textbf{0.96299}                   & 2000                    \\
Town3                      & X                           & 3.97905                           & \textbf{1.09182}                   & 2850                    \\
Town4                      & 35.77386                    & 21.24930                          & \textbf{1.06044}                   & 3000                    \\
Town5                      & X                           & 18.59390                          & \textbf{1.13715}                   & 1000                    \\ \hline
\multicolumn{3}{l}{*The bolded values represent the optimal values for each line.}
\end{tabular}
}
\renewcommand{\arraystretch}{1}
\end{table}
\begin{figure}
    \centering
    \includegraphics[width=0.9\linewidth]{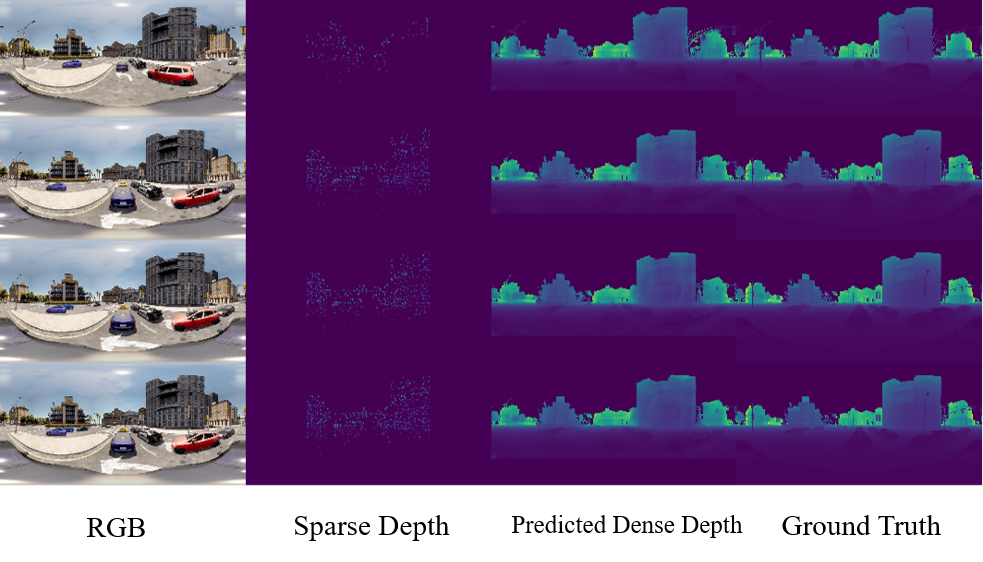}
    \caption{Visualization results for dense depth complementation}
    \label{fig:Depth_compare}
\end{figure}
\subsubsection{Depth Completion}
The real sparse depth points used for training the depth completion network in this study range from 500 to 2000, primarily consisting of textured feature points in the images. During the dense depth acquisition stage, the map points recovered by the panoramic triangulation module are used as sparse depth input and combined with RGB images to reconstruct the dense depth map. Fig.~\ref{fig:Depth_compare} illustrates the qualitative results obtained after applying the multimodal fusion-based depth completion method. From the visual results, it is evident that despite the sparse depth input containing a very limited number of data points, the dense depth maps recovered in this study exhibit high accuracy and closely approximate the true values in various scenes.
\begin{figure}
    \centering
    \includegraphics[width=1\linewidth]{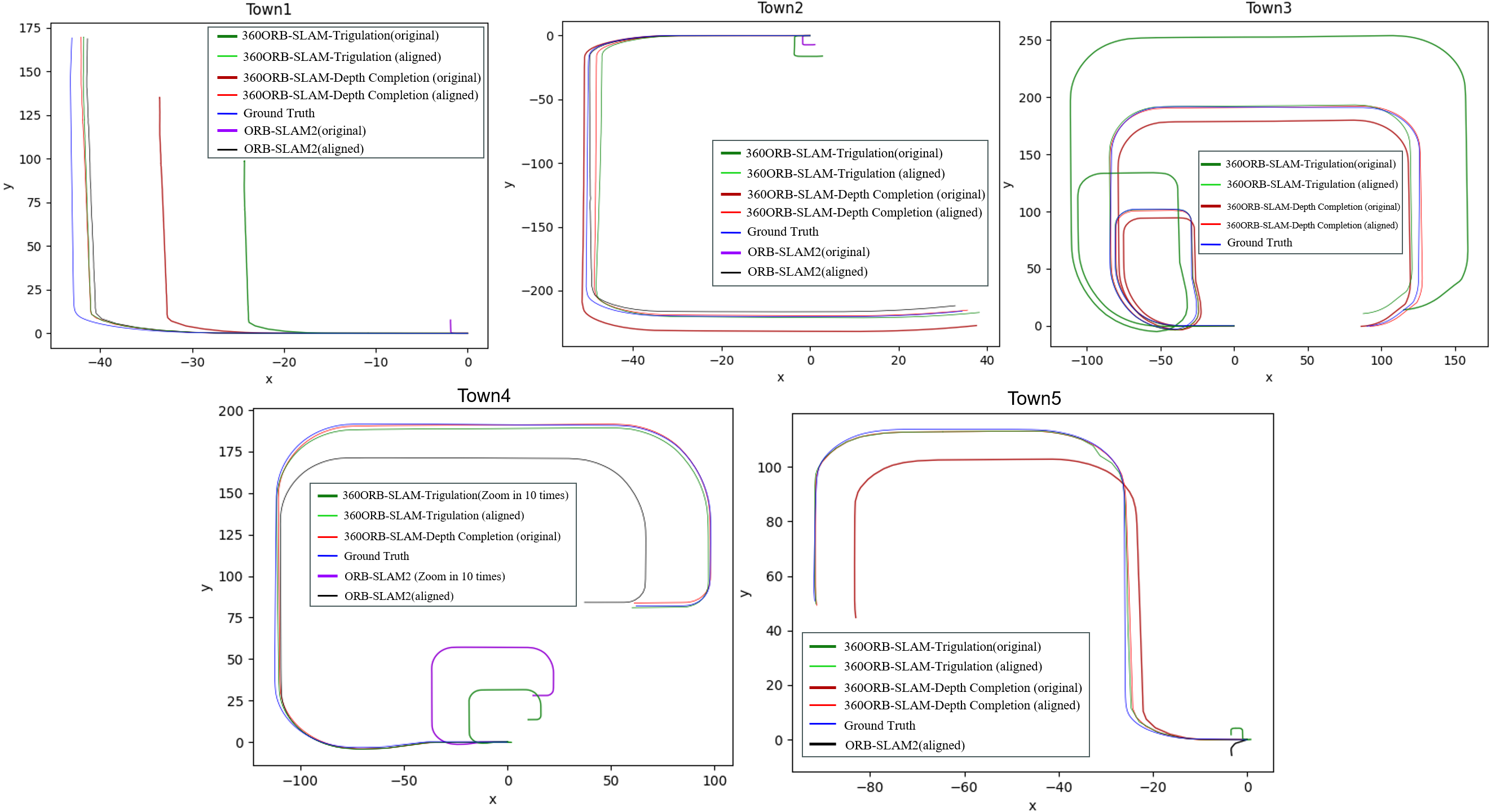}
    \caption{Comparison of the estimated trajectories of the sequences of the dataset with the ground truth}
    \label{fig:Tra}
\end{figure}
\subsubsection{Localization and Mapping}
In this paper, the camera poses were computed for all keyframes, and they were aligned with the ground truth trajectory to measure the RMSE of ATE between the estimated trajectory and ground truth trajectory, as well as the SF between predicted and ground truth trajectories. The quantitative experimental results are presented in Table II and Table III. It can be observed that our system achieved significant improvements in scale accuracy through the dense depth completion method, effectively addressing the scale ambiguity issue in monocular cameras. Regarding the ATE, although the proposed method slightly lags behind the state-of-the-art methods in the Town1, Town2, and Town4 sequences, it is important to note that the trajectory lengths for each sequence are several kilometers long, while the measurement errors are only in the centimeter range and show no significant differences. This is because the proposed method, in the process of dense depth completion, optimizes for both scale and accuracy errors, which affects the precision of trajectory estimation.

Fig.~\ref{fig:Tra} further illustrates the visual comparison between predicted and ground truth trajectories for each sequence. In this work, a scale alignment process was applied to align the scale of some trajectories that had excessively small scales, resulting in improved visual contrast. For example, in the Town4 sequence, the original sizes of 360ORB-SLAM-Tri and ORB-SLAM2 were scaled up by a factor of 10. By comparing different trajectories, it becomes evident that the proposed depth completion method, which provides scale-accurate depth information, outperforms ORB-SLAM2 and 360ORB-SLAM-Tri in terms of scale accuracy in mapping and localization tasks.

\begin{figure}
    \centering
    \includegraphics[width=1\linewidth]{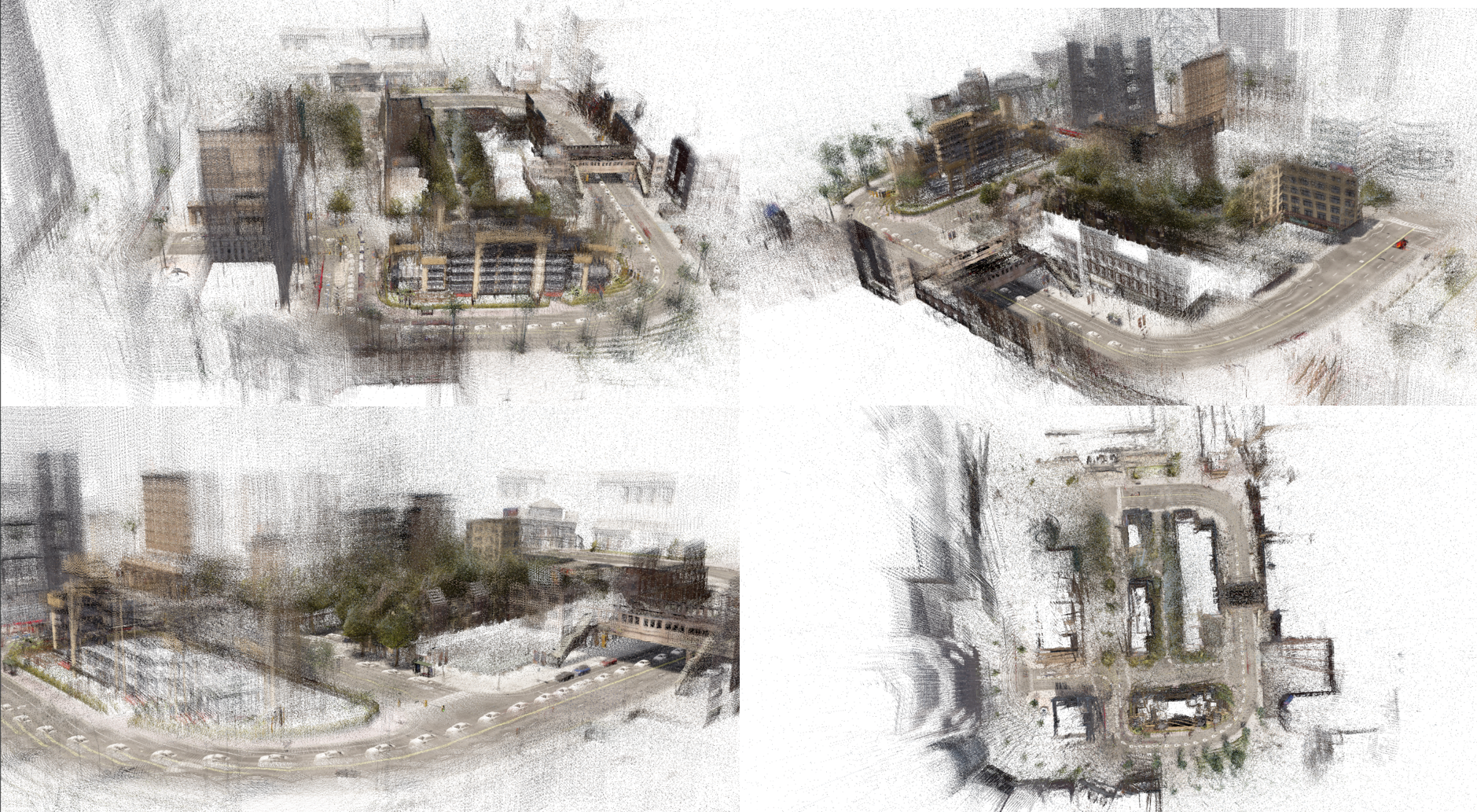}
    \caption{Dense 3D pointcloud reconstruction results for different viewpoints}
    \label{fig:pointCloud}
\end{figure}

Fig.~\ref{fig:pointCloud} illustrates a schematic representation of the dense 3D point cloud reconstructed based on the poses and the 3D map points obtained through the panoramic triangulation process. The dense 3D point cloud provides high-precision spatial information about the environment, which serves as a crucial foundation for navigation, obstacle avoidance, and 3D modeling purposes.
\section{Conclusion}
In this paper, we propose a panoramic camera-based visual SLAM system to address the limitations of traditional feature-based methods like ORB-SLAM in effectively using image information and resolving scale ambiguity in monocular cameras. The system incorporates a panoramic triangulation module for improved map point accuracy, a depth completion network for enhanced depth estimation, and applies deep learning-derived dense depth maps to mitigate scale drift and improve pose estimation accuracy. Experimental results reveal significant performance improvements over ORB-SLAM2, demonstrating the effectiveness of our approach in panoramic image SLAM applications, with potential for real-world implementation and further optimization in future research.

\section*{Acknowledgement}
The authors acknowledge the Supercomputing Center of Hangzhou Dianzi University for providing computing resources.

\end{document}